\documentclass[letterpaper,10pt,conference]{IEEEtran}

\usepackage{copyright}

\usepackage[english]{babel}

\usepackage{amssymb}
\usepackage{amsmath}
\usepackage{hyperref}
\usepackage{cleveref}
\usepackage{graphicx}
\usepackage{latexsym}
\usepackage{algorithm}
\usepackage{algpseudocode}
\usepackage{todonotes}
\usepackage{newclude}
\usepackage{multicol}
\usepackage{tabularx}
\usepackage{booktabs}
\usepackage{threeparttable}
\let\labelindent\relax
\usepackage{enumitem}
\usepackage{listings}
\usepackage{subfigure}
\usepackage{combelow}
\usepackage{cite}

\usepackage{url}
\newcommand\rurl[1]{%
  \href{http://#1}{\nolinkurl{#1}}%
}

\usepackage[binary-units]{siunitx}
\usepackage[acronym]{glossaries}
\newacronym{wsl}{WSL}{Weakly-Supervised Learning}
\newacronym{fmcw}{FMCW}{Frequency-Modulated Continuous-Wave}
\newacronym{mmw}{MMW}{Millimetre-Wave}
\newacronym{ml}{ML}{Machine Learning}
\newacronym{dl}{DL}{Deep Learning}
\newacronym{fcnn}{FCNN}{Fully Convolutional Neural Network}
\newacronym{fov}{FOV}{Field-of-view}
\newacronym{lidar}{LiDAR}{Light Detection and Ranging}
\newacronym{tof}{TOF}{time-of-flight}
\newacronym{sdof}{6DoF}{Six degree-of-freedom}
\newacronym{slerp}{SLERP}{Spherical Linear Interpolation}
\newacronym{ro}{RO}{Radar Odometry}
\newacronym{fabmap}{FAB-MAP}{Probabilistic Localisation and Mapping in the Space of Appearance}
\newacronym{gps}{GPS}{Global Positioning System}
\newacronym{vo}{VO}{Visual Odometry}
\newacronym{slam}{SLAM}{Simultaneous Localization and Mapping}

\definecolor{CommentPran}{rgb}{0.2,0.8,0.2}
\definecolor{CommentDani}{rgb}{0,0,1}
\definecolor{CommentMatt}{rgb}{1,0.2,0}
\definecolor{CommentPaul}{rgb}{0.9,0,0}
\definecolor{CommentReview}{rgb}{0.9,0.6,0.2}

\IEEEoverridecommandlockouts
\IEEEoverridecommandlockouts

\crefname{table}{Table}{Tables}
\crefname{figure}{Figure}{Figures}
\crefname{section}{Section}{Sections}

\renewcommand{\baselinestretch}{0.976}

\title{RSS-Net: Weakly-Supervised Multi-Class \\Semantic Segmentation with FMCW Radar}
\author{Prannay Kaul, Daniele De Martini, Matthew Gadd, Paul Newman\\Oxford Robotics Institute, Dept. Engineering Science, University of Oxford, UK.\\\texttt{\{prannay,daniele,mattgadd,pnewman\}@robots.ox.ac.uk}}

\begin{document}

\maketitle

\begin{abstract}
This paper presents an efficient annotation procedure and an application thereof to end-to-end, rich semantic segmentation of the sensed environment using \acrlong{fmcw} scanning radar.
We advocate radar over the traditional sensors used for this task as it operates at longer ranges and is substantially more robust to adverse weather and illumination conditions.
We avoid laborious manual labelling by exploiting the largest radar-focused urban autonomy dataset collected to date, correlating radar scans with RGB cameras and \acrshort{lidar} sensors, for which semantic segmentation is an already consolidated procedure.
The training procedure leverages a state-of-the-art natural image segmentation system which is publicly available and as such, in contrast to previous approaches, allows for the production of copious labels for the radar stream by incorporating four camera and two \acrshort{lidar} streams.
Additionally, the losses are computed taking into account labels to the radar sensor horizon by accumulating \acrshort{lidar} returns along a pose-chain ahead and behind of the current vehicle position.
Finally, we present the network with multi-channel radar scan inputs in order to deal with ephemeral and dynamic scene objects.
\end{abstract}


\glsresetall

\copyrightnotice

\section{Introduction}%
\label{sec:introduction}

Safe navigation and operation of mobile robots in search and rescue, agriculture, and mining environments will require perception systems that deliver a detailed understanding of the surroundings regardless of adverse environmental factors.

\gls{lidar} and vision-based systems have been widely investigated and adopted in the last decade.
However, these models are usually trained on datasets such as~\cite{Cordts2016Cityscapes} which are captured in uniform conditions, leaving the state-of-the-art susceptible to rain, snow, fog, glare, lighting, and seasonal appearance changes.

\begin{figure}
    \centering
    \includegraphics[width=\columnwidth]{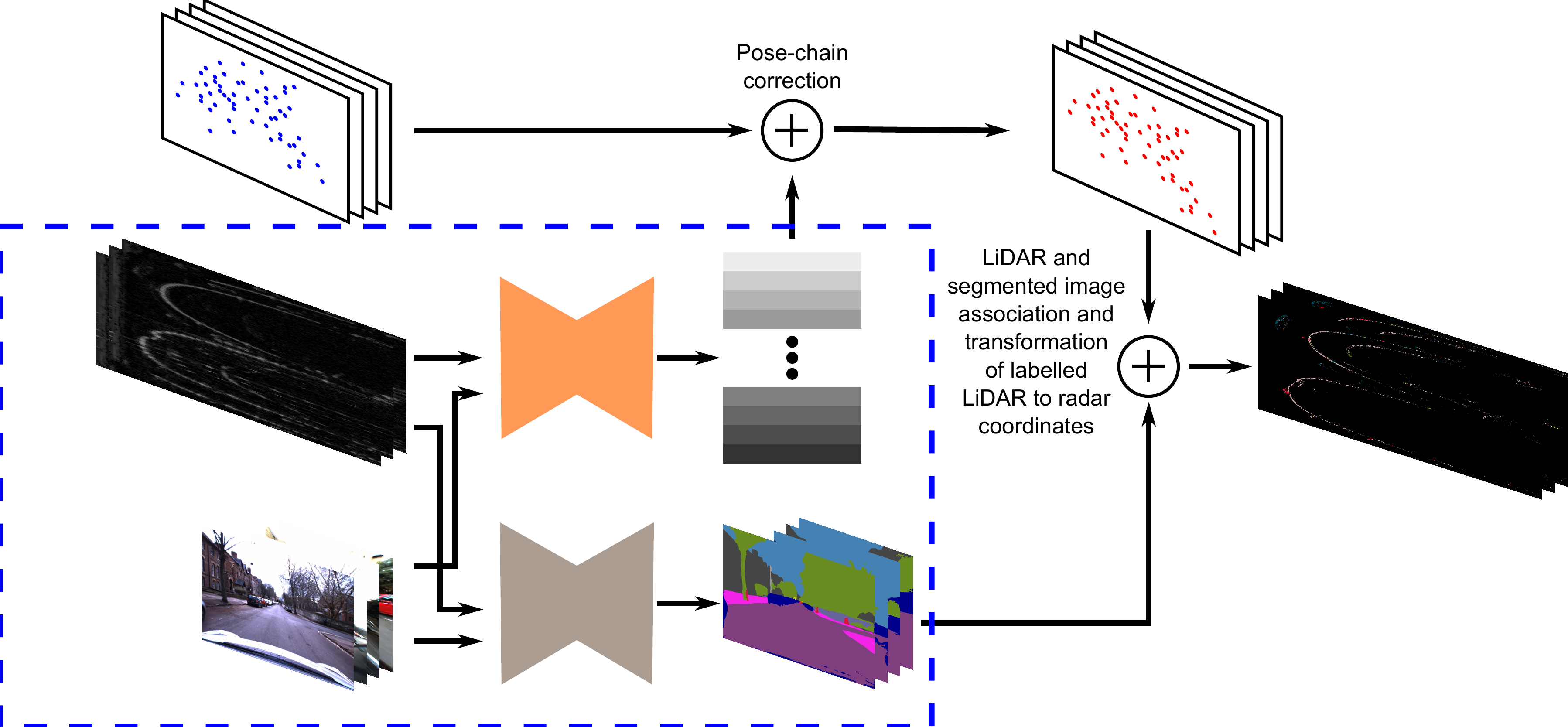}
    \caption{An overview of the pipeline implemented to generate labelled training data for radar segmentation. The section within the blue box is completed before training such that segmented RGB streams are available on disk, as is the pose chain described in~\cref{ssec:posechain}.
    During training, a radar scan is selected from the training set.
    The temporally nearest RGB images and corresponding \gls{lidar} scans are then used to form the labelled radar image as described in~\cref{sec:train_data_gen}.
    The resulting data are therefore formed on the fly during the training/testing process.}
    \label{fig:pipeline}
\end{figure}

\begin{figure*}[t]
    \centering
    \subfigure{\includegraphics[width=0.85\textwidth]{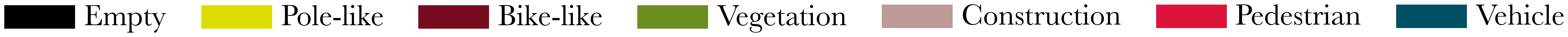}}\\
    \subfigure[]{\includegraphics[width=0.28\textwidth]{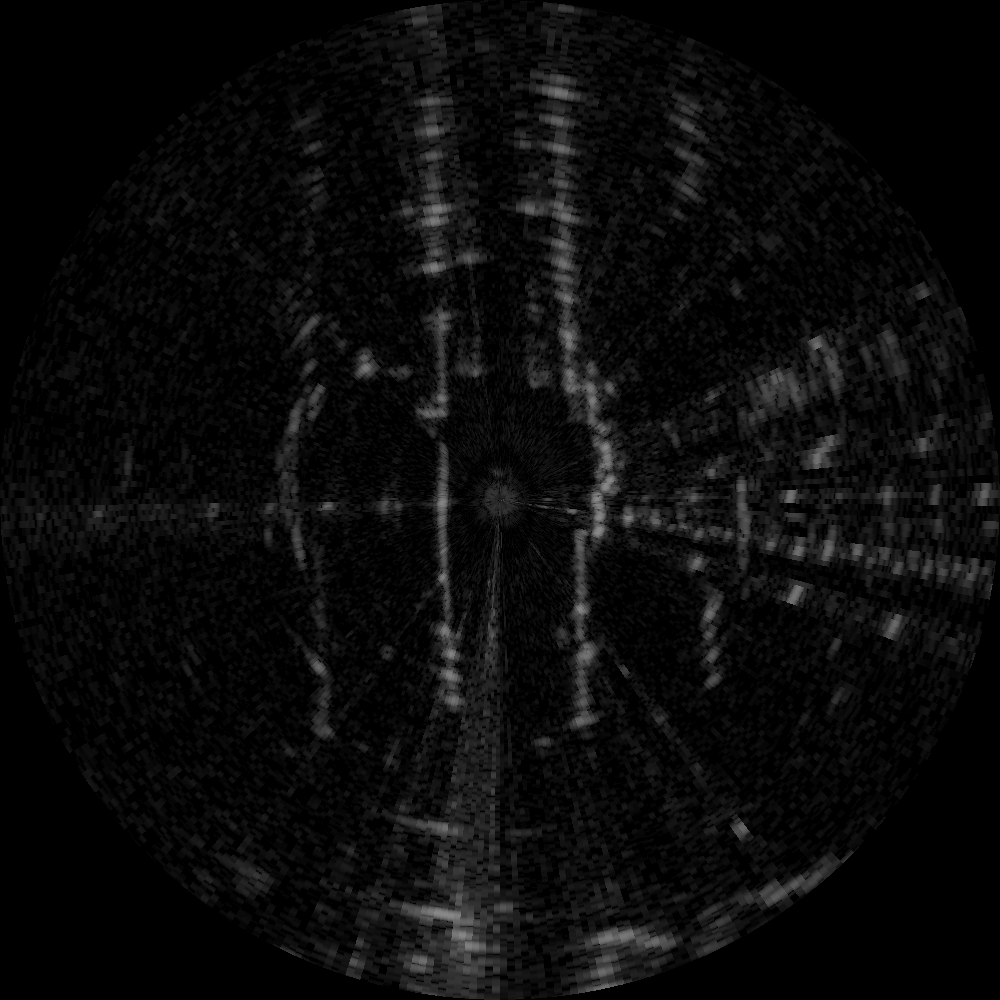}\label{fig:cart_input}}
    \subfigure[]{\includegraphics[width=0.28\textwidth]{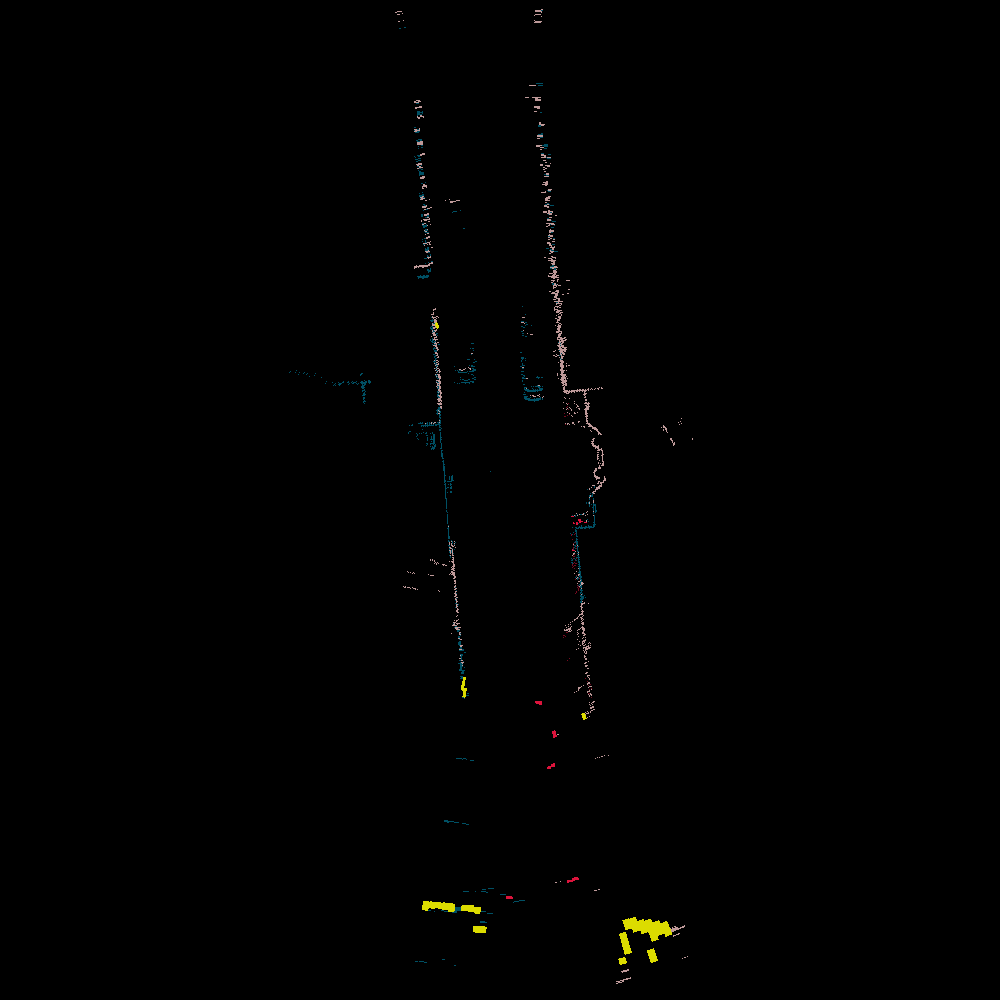}\label{fig:cart_target}}
    \subfigure[]{\includegraphics[width=0.28\textwidth]{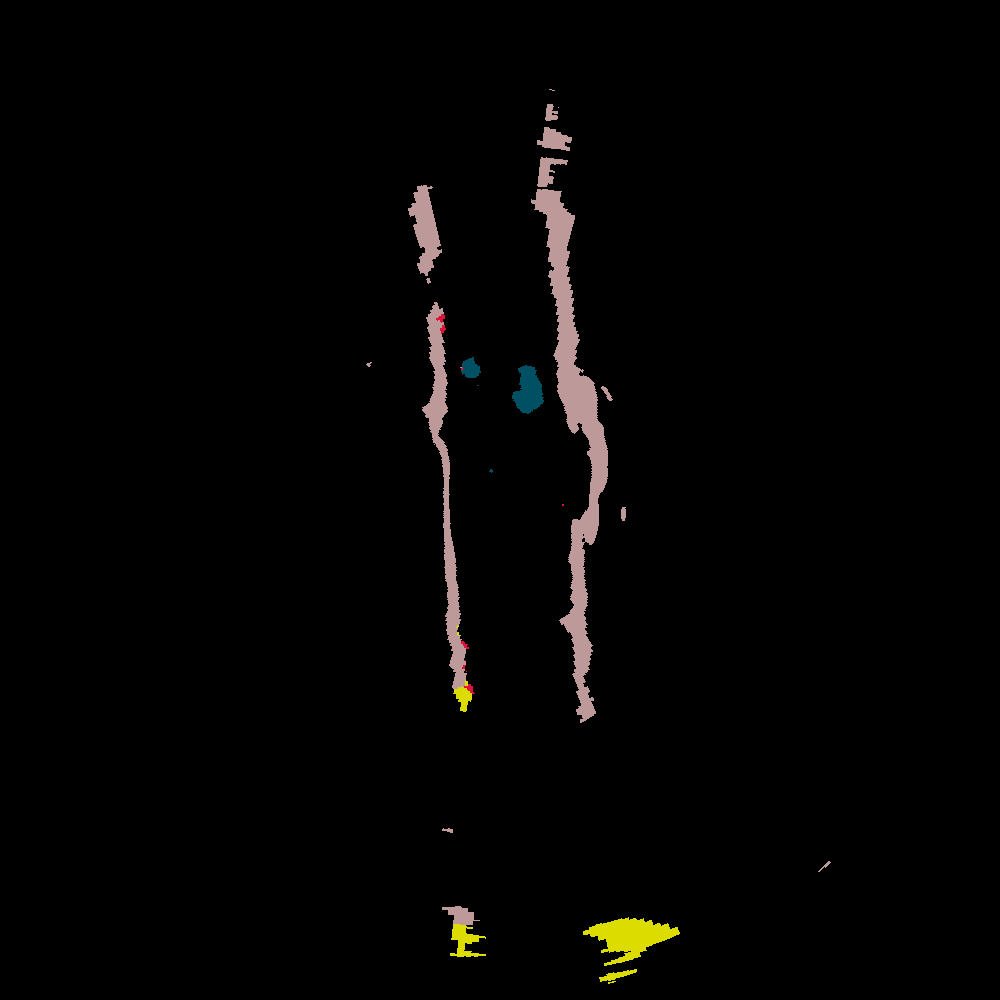}\label{fig:cart_pred}}
    \caption{Comparison of a \subref{fig:cart_input} raw radar scan, \subref{fig:cart_target} labelled targets for training the network, and \subref{fig:cart_pred} predicted semantic segmentation, all shown in Cartesian representation.
    Here, the Cartesian representation takes the form of an image with resolution $N{\times}N = 1000{\times}1000$, where $N$ is the cardinality of the set of bins in the discretised range returns of the original polar representation.}
    \label{fig:images_radar}
\end{figure*}

In contrast, due to the long wavelength of radio waves, \gls{fmcw} scanning radar operates well under variable weather and lighting conditions.
Additionally, multiple returns are received from a single azimuthal transmission and operation at ranges of up to hundreds of metres is common.
Indeed, there is a burgeoning interest in exploiting \gls{fmcw} radar to enable robust mobile autonomy, including ego-motion estimation~\cite{cen2018precise,cen2019radar,aldera2019,2019ITSC_aldera,2019_Barnes_Arxiv_MaskingByMoving} and localisation~\cite{KidnappedRadarArXiv,tang2020rsl}.
However, the radar measurement process is complex and scans formed thereby are prone to pollution by multipath reflections, speckle noise, and other artefacts in addition to the internal noise characteristics~\cite{adams2012robotic}.
Furthermore, radar scans compress the 3D environment into a 2D planar representation, which limits the discrimination between objects which have similar appearance.

\cref{fig:images_radar} shows an urban scene as perceived by a \gls{fmcw} radar and the corresponding segmentation from the proposed system.
\cref{sec:network} presents a method to segment this raw radar scan using a \gls{fcnn}.
This system is designed with the principle that sensor artefacts, which may appear similar to actual objects, are to be ignored and objects with distinct typical dynamics should be reliably and unambiguously identified.
As even radar experts find labelling non-intuitive,~\cref{sec:train_data_gen} proposes a novel annotation technique suitable for weak supervision using alternative sensing modalities also present on the mobile robot (RGB and \gls{lidar} streams), leveraging a state-of-the-art, publicly available, pre-trained model.

\section{Related Work}%
\label{sec:related}

The success of \gls{dl} in general perception tasks, such as object classification and semantic segmentation of natural images, is well documented in the community~\cite{agarwal2018}, benefits of which are now impacting radar sensing~\cite{2019_Barnes_Arxiv_MaskingByMoving,KidnappedRadarArXiv,tang2020rsl}.

Occupancy grid methods~\cite{weston2018probably} provide discrimination between free and occupied space, dealing with the complex sensor artefacts encountered in radar scans, but fail to deliver richer information regarding the nature of the occupied space.
To overcome this, a \gls{fcnn} is used in~\cite{lombacher2017semantic} to segment a radar scan based on the probability of occupancy and hand labels are used to classify objects.
However, hand labelling of radar scans is a time-consuming task; therefore, as a trade-off between quantity and quality of labelled data, we use a form of weak supervision: semantically segmenting camera streams and combining the result with \gls{lidar} range measurements to provide a labelled image in the radar frame.

Feature-based clustering approaches have been shown to work in~\cite{scheiner2018radar}, while in~\cite{scheiner2019radar} an ensemble, where each classifier takes as input its own specialised feature set, allows for unseen patterns to be detected.
In contrast to~\cite{scheiner2018radar,scheiner2019radar} and similarly to~\cite{schumann2018semantic} our work does not use hand-crafted features nor ensembles of classifiers.
Furthermore, in contrast to~\cite{schumann2018semantic} we do not use deep networks designed for pointclouds and instead use popular image segmentation networks on the the radar scans.
Moreover, in contrast to all of~\cite{scheiner2018radar,scheiner2019radar,schumann2018semantic} our work does not require manual labelling.

From the definitions in~\cite{zhou2017brief}, our method falls in both \textit{inexact} -- as the radar data and \gls{lidar} data have different granularities and range -- and \textit{inaccurate} -- as the labels are not created by a human supervisor.

An example of both inexact and inaccurate supervision applied to radar is~\cite{weston2018probably}, where the authors use a U-Net, the popular image segmentation network~\cite{ronneberger2015u}, to segment a radar scan based on probability of occupancy on a more extensive dataset and produces good results.
In contrast to~\cite{weston2018probably}, however, where the shorter maximum range of the \gls{lidar} is not dealt with, we accumulate \gls{lidar} returns in the native radar representation using a good external source of ego-motion.

A second example of inaccurate supervision can be found in~\cite{aldera2019}: an automatic labelling procedure is carried on to classify readings in radar scans as consistent across wide baselines or not; a U-Net is then trained to predict pixels in the polar representation of the radar scan.
The result is a filtering technique, which limits the number of landmarks for the odometry pipeline, making the pipeline faster without losing accuracy.

Finally, in the same spirit as us, the authors of~\cite{hoermann2018dynamic} apply a \gls{wsl} approach to train a network for object detection on dynamic grid maps.
They exploit temporal and spatial relationships to extract moving objects and their shapes using a \gls{lidar} sensor.
Indeed, once an object is observed, it continues being observed until it exits the field of view, updating shape and trajectory information.
Then, this information is propagated backwards in time to refine the annotation for more consistency in the labels.

\section{Training Data Generation}
\label{sec:train_data_gen}

\cref{fig:pipeline} provides an overview of the pipeline used to generate training data for each of the radar scans.
We leverage the depth of research by the computer vision community in the semantic segmentation of RGB images and specifically of urban street scenes.
The offline datasets we use contain concurrent RGB image streams covering the full azimuth range (four cameras with \SI{360}{\degree} horizontal field-of-view), \gls{lidar} scans from two lasers, and radar scans.

\subsection{Semantic Image Segmentation}\label{ssec:SemanticImageSeg}

In order to create labelled images for the training procedure, we use the publicly available Deeplabv3-DPC~\cite{chen2018searching} model, trained on the Cityscapes dataset, to perform semantic image segmentation on monocular visual image streams, identifying each pixel as belonging to one of a number of meaningful object categories.
The Cityscapes dataset is comprised of forward facing urban street scenes, in contrast to our RGB streams which contain camera streams in four directions (forward, rear, left, and right).
Furthermore, each stream is captured during dynamic motion and so many images are impacted by motion blur (particularly near the edges) and exposure effects.

\subsection{LiDAR and RGB Image Fusion}
\label{ssec:LiDARImageFus}

Given the rich semantic segmentation extracted from the RGB streams, a method is required to use this information to generate the labelled radar training data. Using extrinsic transformations between the seven sensors (four RGB cameras, two \gls{lidar} scanners and one radar) and the intrinsic parameters of the four RGB cameras, one can project the \gls{lidar} pointcloud at a given time-step onto each corresponding camera image.
By representing the pointcloud in image coordinates a label for each point within the image can be extracted using the segmented images.
After associating a label with each point of the \gls{lidar} scan, using each of the four camera streams, the extrinsic transformation between the relevant \gls{lidar} scanner and the radar is used to form labelled data in the radar coordinate frame.
Finally these points are projected into the horizontal plane and discretized into range-azimuth grid cells.
For the case in which multiple labelled points map to the same pixel location, the label is selected with equal probability.

\subsection{Pose-chain Interpolation of Labels}
\label{ssec:posechain}

The \gls{lidar} scans from each sensor are gathered at \SI{20}{\hertz}, whereas the camera and radar streams are gathered at \SI{25}{\hertz} and \SI{4}{\hertz}, respectively.
Furthermore there is a non-negligible difference between the times at which the radar scan is taken and the temporally closest four images (one image for each direction) and \gls{lidar} scan.
Due to this temporal difference and the dynamic nature of the environment which the sensors are operating within, the projection of the pointcloud into a given camera image suffers from misalignment.

We correct for this misalignment through interpolating an optimised pose chain between the time of \gls{lidar} capture and its temporally closest images, one from each camera.
Given this pose-chain, the required transformation is obtained through interpolation.
The rotational component is obtained by \gls{slerp} on a spherical surface traced by a unit quaternion, as described in more detail in~\cite{shoemake1985animating}, and the translational component is obtained by a constant velocity interpolation.

In the same fashion, each radar scan is related to the closest \gls{lidar} pointclouds in time.
Indeed, for each azimuth message, a pointcloud is selected from each sensor ad projected into the scan though motion interpolation and extrinsic transformations.

\subsection{Accumulation of labels to the radar sensor horizon}

As seen in \cref{fig:cart_target}, \gls{lidar} scans are an inherently sparse representation of an environment.
Furthermore, as the datasets used are collected in urban environments, the result is the majority of pointcloud readings are within a relatively short range compared to the full operating range of a \gls{fmcw} radar and very few meaningful labels available at a distance past \SI{40}{\metre}.
Already having access to the pose-chain described in \cref{ssec:posechain}, we employ it once again to transform labelled \gls{lidar} pointclouds before and after the current radar scan along the trajectory of the robot into the radar frame.

\begin{figure}
    \centering
    \subfigure[]{\includegraphics[width=0.49\columnwidth]{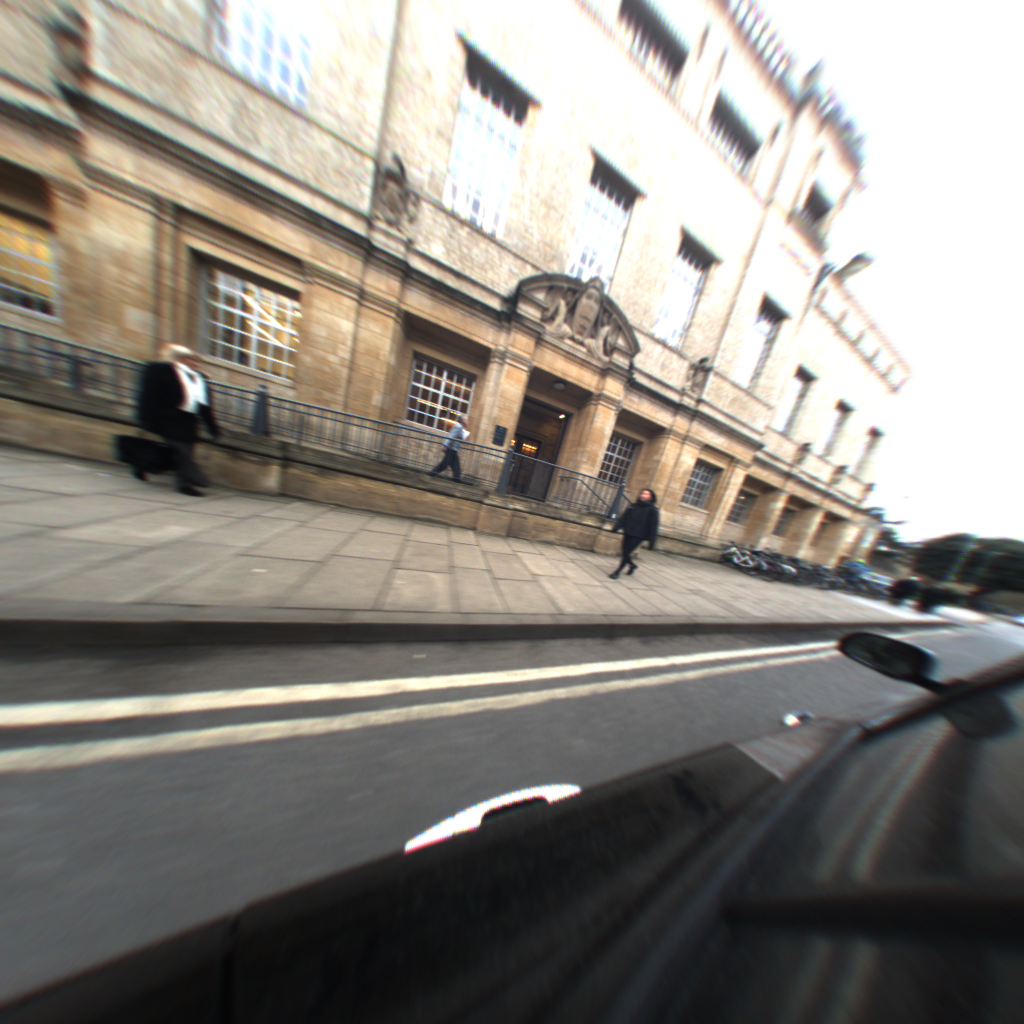}}
    \subfigure[]{\includegraphics[width=0.49\columnwidth]{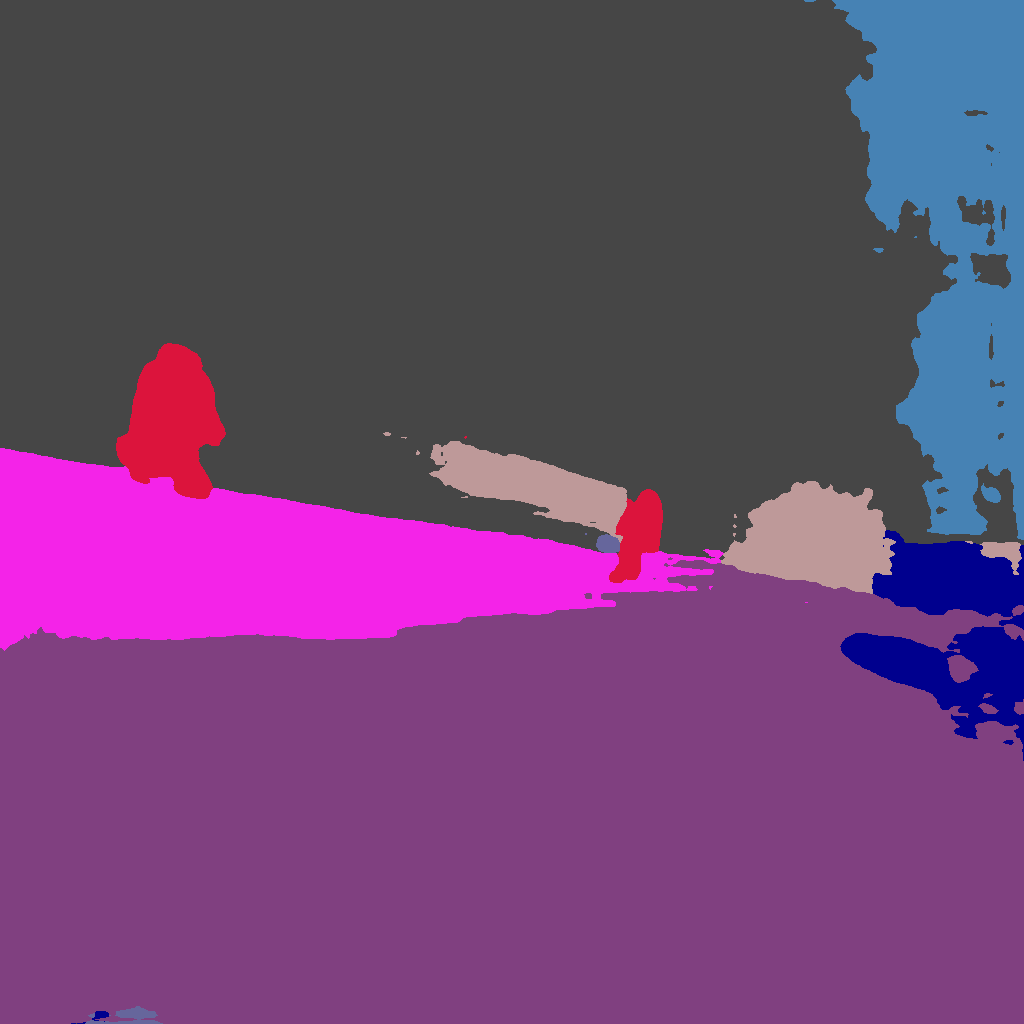}}
    \caption{Example of \emph{one of} the four RGB streams, extracted at the same time-step, with the corresponding segmentation extracted from Deeplabv3-DPC.
    In this example, taken from one of the wide-angle monocular cameras, one observe motion blur, non-ideal exposure and a reduction in segmentation quality as the edge of the image is approached.}
    \label{fig:images_rgb}
\end{figure}

\section{Network Architecture for Radar Semantic Segmentation}\label{sec:network}

\begin{figure*}
    \centering
    \includegraphics[width=0.75\textwidth]{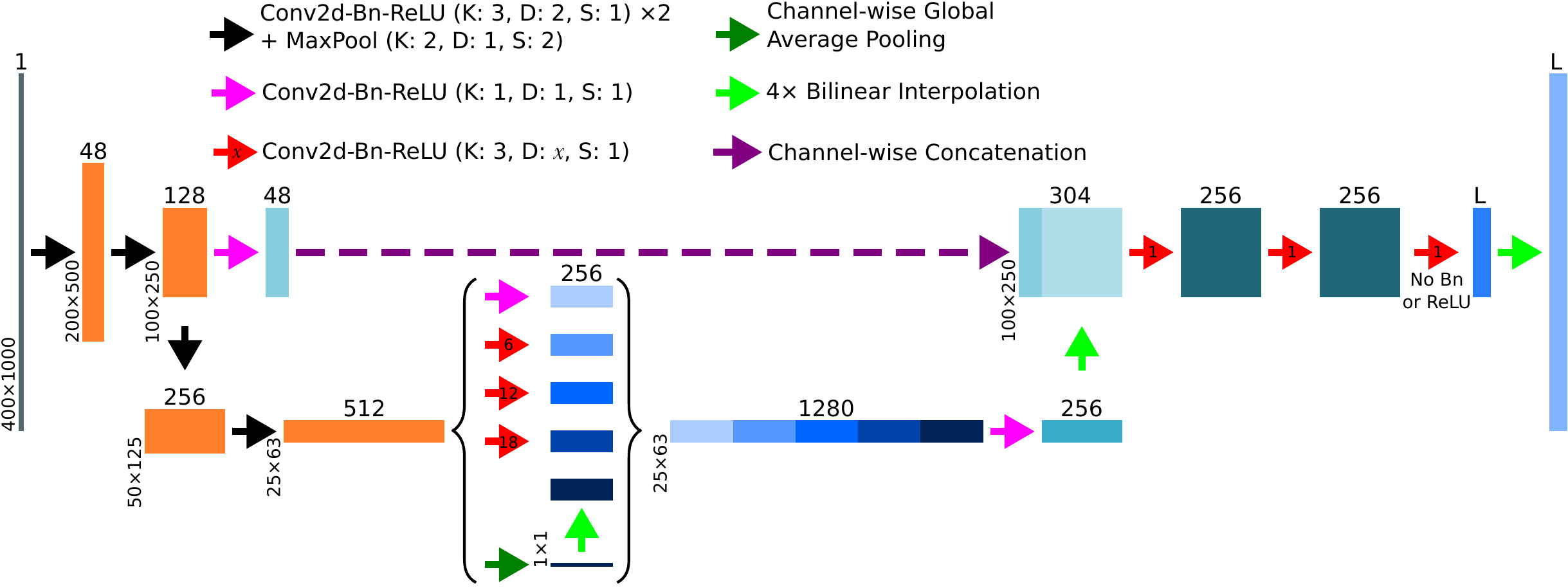}
    \caption{Overview of the network used for radar semantic segmentation. L--number of classes. Bn--2D batch normalization, ReLU--Rectified Linear Unit, K--kernel size (all kernels are square), D--dilation rate, S--convolution stride.}
    \label{fig:net_architecture}
\end{figure*}

\cref{fig:net_architecture} provides an illustration of the \gls{fcnn} architecture we employ. The design makes use of dilated convolutions initially, in the encoder section (orange channels) to increase the receptive field exponentially. The sparse nature of the generated training data requires fine details to be managed by the network, therefore only 4 max-pooling layers are used in total to prevent a high loss of detail from the input image, making the smallest feature map have dimensions sixteen times smaller than the input. The rich features after four sections of encoding are passed through an atrous (dilated) spatial pyramid pooling (ASPP) block introduced in~\cite{chen2018encoder}, which uses varying rates of dilated convolution and global average pooling to produce rich semantic information at various scales. The output feature maps of these different dilated convolutions are concatenated together and reduced to $256$ channels using $1\times1$ convolutions. The bilinear upsampled versions of these feature maps are concatenated with higher resolution feature maps from earlier in the network. This provides a mix between rich semantic information and fine detailed feature maps. Subsequent convolutions are used to reduce $304$ feature maps to $L$ channels which are bilinearly upsampled to give a final output with size equal to the input. During training, the $L$ output channels are passed into a cross entropy loss function along with the true labels (generated as described in~\cref{sec:train_data_gen}). During testing, the $L$ output channels are passed through an `argmax' layer yielding a single channel output with class labels assigned to each element/pixel. In the above explanation, $L$ is the number of classes segmented by the network.

\subsection{Class Weighting}\label{ssec:class_weighting}
As the target images are generated from \gls{lidar} scans which are inherently sparse, there is a large class imbalance in the dataset.
To overcome this, the loss function is weighted depending on the true class using a logarithmic function:

\begin{align}
w[i] \propto \left(1 + \log\frac{\sum^{N}_{j=1} t[j]}{N \cdot t[i]}\right)^2
\end{align}
where $w[i]$ is the class weight, $t[i]$ is the number of pixels belonging to class $i$ in the training set and $N$ is the number of classes. It should be noted that the weight for Empty class (see~\cref{ssec:class_definitions}) is empirically set to $0.1$.

\subsection{Addressing Dynamic Objects}\label{ssec:dynamic}
Different classes in a single radar scan often appear near-identical (e.g. a ``Pedestrian'' and ``Pole-like'') even to an expert, they exhibit spatial similarity.
The discriminating feature in such cases is the dynamics of the object -- a ``Pole'' is stationary, but a ``Pedestrian'' is (usually) moving -- they exhibit temporal dissimilarity.
To improve performance in such cases, three consecutive radar scans are input to the \gls{fcnn}, providing the network with temporal information.
Pose-chain interpolation is used to transform the scans into the same frame of reference; this ensures that stationary objects appear in the same location across all three scans.
Given the pose-chain is in Cartesian coordinates, the radar scans are input to the network in Cartesian form.

Despite adding a temporal component the network remains fully convolutional and no recurrent units are used.
To aid the network to learn useful distinct temporal and spatial kernel weights, depthwise separable convolutions are used throughout the network.
Depthwise separable convolutions provide the added benefit of reduced network parameters and increased training time~\cite{howard2017mobilenets}.

\section{Experimental Setup}%
\label{sec:application}

The experiments are performed using data collected from the \textit{Oxford RobotCar} platform~\cite{maddern20171}.

\subsection{Ground truth radar ego-motion}

We use the ground truth pose dataset described in the recently released \textit{Oxford Radar RobotCar Dataset}~\cite{OxfordRadarRobotCarDataset} which is computed by an optimisation using \gls{gps}, robust \gls{vo}~\cite{barnes2018driven}, and visual loop closures from FAB-MAP~\cite{cummins2008fab}.

\subsection{Dataset Demarcation for Training and Validation}
\label{ssec:Demarcation}

The route in Oxford city centre used for collecting the offline datasets is taken from~\cite{tang2020rsl}.
The path has been divided into three different portions: \textit{train} (blue), \textit{validation} (black) and \textit{test} (purple).
We specifically designed the sets in such a way that no intersection would occur among the three of them.
Although the radar scanner can sense up to hundreds of metres, the environment taken in consideration is cluttered and does not allow it to reach such sensing ranges; thus we decided not to discard any sample on the dataset, but only to add a little padding between the portions, of the order of ten metres.
The resulting dataset is then formed by \SI{6246}{} training, \SI{314}{} validation and \SI{1720}{} test examples.

\begin{figure*}
    \centering
    \subfigure[]{\includegraphics[width=\columnwidth]{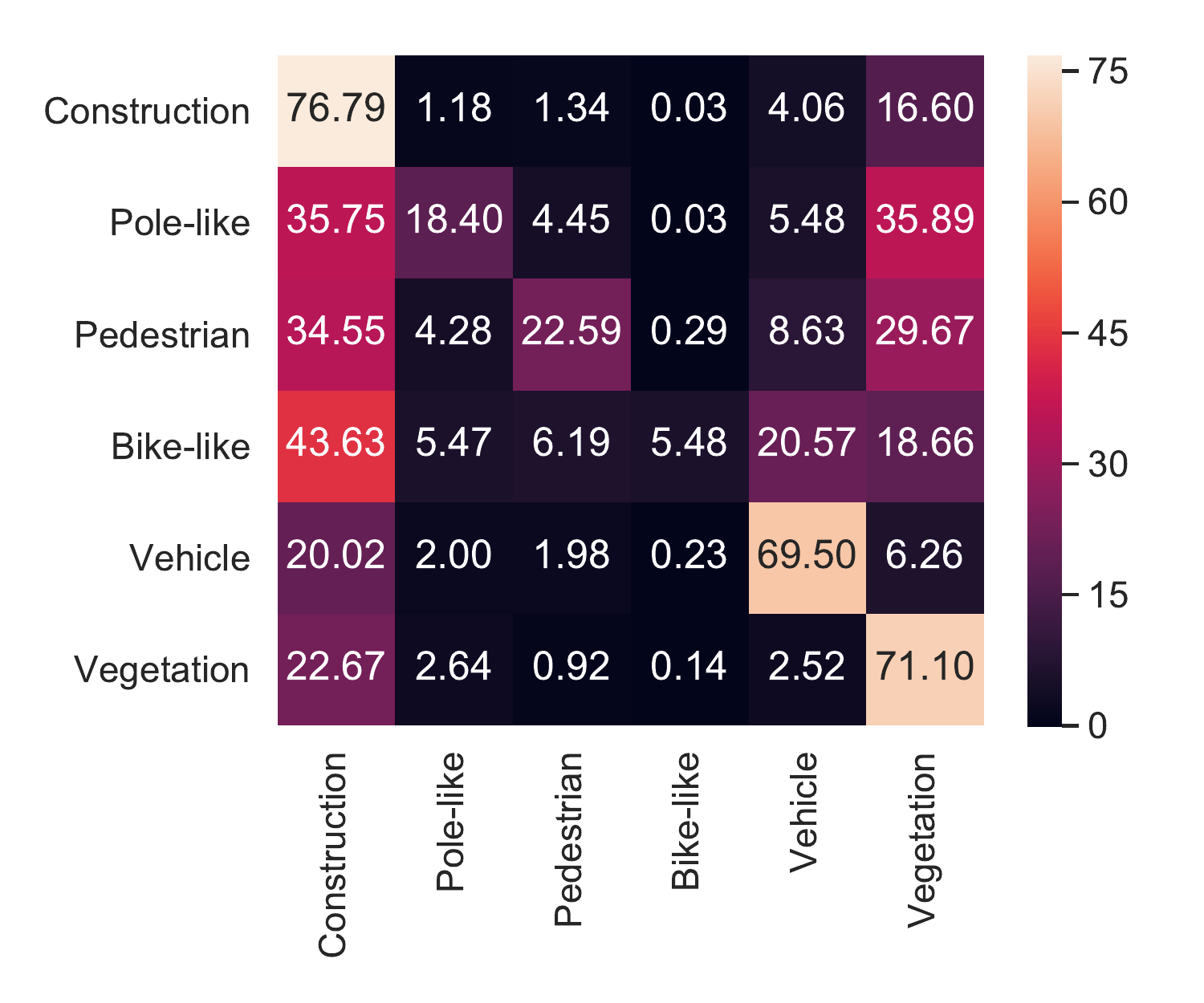}\label{figs:confusion_full}}
    \subfigure[]{\includegraphics[width=\columnwidth]{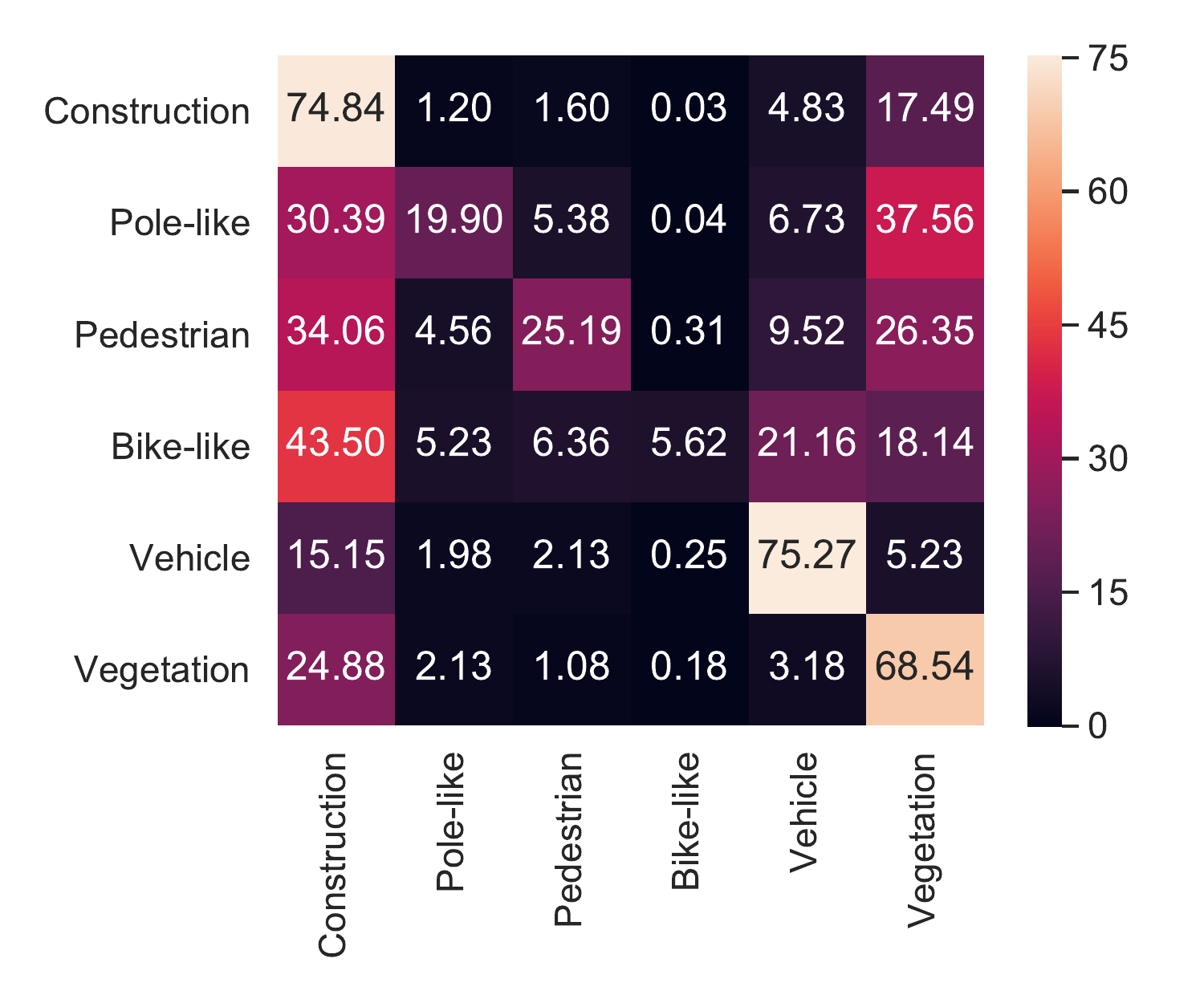}\label{figs:confusion_short}}
    \caption{Normalized confusion matrix on the test set considering \subref{figs:confusion_full} the full radar sensing horizon and \subref{figs:confusion_short} a foreshortened radar sensing horizon.}
    \label{fig:confusion_matrix}
\end{figure*}

\subsection{Data Augmentation}

Data augmentation has been taken into account in the training phase to increase the number of training examples. We simply add a random flips on the horizontal and vertical axis with a $50$\% probability.

\subsection{Class Definitions}\label{ssec:class_definitions}
The Cityscapes dataset uses $34$ distinct classes with $19$ recommended for use during training.
Use of such a rich class dictionary is not suitable for radar scans.
For example, the Cityscapes dataset makes distinction between buildings, walls and fences.
The difference between these classes are rarely based on dimension, but instead on appearance and so in a radar scan they appear nearly identical.
A similar case can be made for traffic signs, street lights and poles.
Furthermore, certain classes are unobservable in radar scans, such as roads, sky and grass areas.
For these reasons certain classes are omitted when generating the labelled training data and others are grouped together.
Although the road is not observable, it can be inferred to from the robot location, which in turn could provide more information regarding the presence of sidewalks or grass terrain.
Including such priors is not addressed in this paper and is left for future work.

In total the network uses $7$ object classes. These are grouped in the following way with the labels from the Cityscapes dataset shown in parentheses:
\begin{enumerate}
    \item{Empty (sky, road, sidewalk, terrain, guard rail etc.)}
    \item{Construction (building, wall, fence)}
    \item{Pole-like (poles, traffic lights, traffic sign)}
    \item{Pedestrian (person)}
    \item{Vehicle (car, truck, bus, caravan, trailer, train)}
    \item{Bike-like (rider, bicycle, motorcycle)}
    \item{Vegetation (vegetation)}
    \item[\vspace{\fill}]
\end{enumerate}

Although ``Pole-like'' objects and ``Pedestrians'' generally appear similar in radar scans, we decided to maintain separate classes, since different vehicle behaviour is needed when either poles or people are present.

\section{Results}%
\label{sec:results}

\cref{figs:confusion_full} shows the distributions of predicted and true class memberships. Most classification errors arise from non-Construction objects being labelled as Construction objects.
Moreover,~\cref{figs:confusion_short} shows the confusion matrix when only considering the closest parts of the environment. The relatively small difference in performance demonstrates that the network has generated representations which are equivariant in the polar form of the radar scans and understands the varying appearance of objects with changing radial distance from the robot.

\cref{fig:images_radar} shows an example from the test set (described in~\cref{ssec:Demarcation}).
One can see from the middle row that a large part clearly in the class `Construction' has been incorrectly labelled as the Vehicle class (observe the bottom left portion of the polar image).
This arise due to two reasons; the first is that despite improving the projection of the \gls{lidar} pointcloud into the image streams, using a pose-chain there will still be a discrepancy between the alignment of the image and the pointcloud.
This leads to incorrect labelling of some regions, particularly at the boundaries between objects where the depth of the scene changes most rapidly.
Secondly, the segmentation of the RGB images itself is not perfect and so even if the alignment between images and pointclouds were exact, incorrect or bloated segmentation labels lead to incorrect labelling in our generated target data; this issue is confounded during evaluation as even though the bottom row shows that incorrectly labelled portion of ``Construction'' is predicted correctly by the network, this will count towards the incorrect labelling of the ``Vehicle'' class.

It is not surprising that larger objects such as ``Construction'', ``Vehicle'' and ``Vegetation'' attain higher accuracies.
As these objects are relatively large, the misalignment of the \gls{lidar} scans with the RGB images leads to proportionately fewer incorrect labelled pixels.
This is in contrast to smaller objects (such as a pole) which even a small misalignment can lead to the labelling of Pole-like where Construction exists and vice versa, for example.
If a higher proportion of labels for smaller objects are incorrect, it is natural to expect inferior performance.
A visual observation of the test results (such as in~\cref{fig:images_radar}) demonstrate that many of the labels which appear to be incorrectly labelled as the Construction class are in fact correctly predicted by the network.

A particular reason for the very poor performance of the Bike-like class is a nuance of the dataset used. The data is collected in the Oxford city centre which has a notorious large number of bicycles.
Bicycles make up over $80$\% of the labels in the ``Bike-like'' class.
Inspecting the RGB streams shows a very large number of these bikes fall into one of two categories.
They are either alone and parked right up against objects belonging to the ``Construction'' class (i.e. walls, buildings and fences) or they are in groups on large bike racks.
Furthermore due to the hollow nature of bicycles there are relatively few labels with bicycle labels near walls.
In the first case, the result is that the network treats bicycle labels as noise, which are then ignored during training, and so are simply predicted to be the ``Construction'' class.
In the second case the racks of bicycles appear similar to ``Vegetation'' in radar scans; for these reasons along with the more general ones outlined above, the performance in the ``Bike-like'' class is very poor.

Finally, in order to behave appropriately, mobile robots operating in urban environments are required to distinguish between a ``Pole-like'' object and a ``Pedestrian''.
However from a single radar image, it is impossible, even for a human expert, to distinguish between them.
It is expected then that within the ``small'' object classes, the network will confuse Pedestrians and Pole-Like objects.

\section{Conclusions}%
\label{sec:conclusions}

This work demonstrates a method for producing large amounts of training data for radar segmentation suitable for weak supervision.
This provides a key benefit as it does not require expensive and laborious manual-labelling of radar scans. Furthermore, radar scans contain complex artefacts and so manual labelling would be limited to experts and might nevertheless be difficult.
Unlike previous work on scene understanding using radar scans, no preprocessing of radar data is needed.
The method of labelled data generation made use of publicly available models trained on publicly available datasets to provide the semantic segmentation of the environment observed in our own offline dataset, enabling methods for future work to utilize the depth of work in this related but not identical field.
The relatively small network ($\sim$ $8$M parameters) currently performs well with larger objects and demonstrates robustness to the imperfections of the data generation process.
In the future we plan to retrain and test the system on the all-weather platform described in~\cite{kyberd2019}.

\section*{Acknowledgements}%
\label{sec:acknowledgements}
Prannay Kaul is supported by UK's Engineering and Physical Sciences Research Council (EPSRC) through the Centre for Doctoral Training (CDT) in Autonomous Intelligent Machines and Systems (AIMS) Programme Grant EP/L015897/1.
Daniele De Martini is supported by the UK EPSRC programme grant EP/M019918/1.
Matthew Gadd is supported by Innovate UK under CAV2 -- Stream 1 CRD (DRIVEN).
Paul Newman is supported by EPSRC Programme Grant EP/M019918/1.
The authors would also like to acknowledge the support of the Assuring Autonomy International Programme, a partnership between Lloyd’s Register Foundation and the University of York.

\bibliographystyle{IEEEtran}
\bibliography{biblio}

\end{document}